\documentclass[nohyperref]{article}

\usepackage{microtype}
\usepackage{graphicx}
\usepackage{subfigure}
\usepackage{booktabs} 
\usepackage{multirow}
\usepackage{makecell}
\usepackage{ amssymb }

\usepackage{hyperref}

\usepackage[accepted]{icml2022}

\usepackage{amsmath}
\usepackage{amssymb}
\usepackage{mathtools}
\usepackage{amsthm}
\usepackage{color}
\usepackage[capitalize,noabbrev]{cleveref}

\theoremstyle{plain}
\newtheorem{theorem}{Theorem}[section]
\newtheorem{proposition}[theorem]{Proposition}

\theoremstyle{definition}

\theoremstyle{remark}

\definecolor{green}{rgb}{0,0.5,0} 
\definecolor{red}{rgb}{1.,0,0}

\newcommand{\greenstar}{\textcolor{green}{$\bigstar$}}
\newcommand{\redstar}{\textcolor{red}{$\bigstar$}}

\usepackage[textsize=tiny]{todonotes}

\icmltitlerunning{Deconfounding Neural Similarity}

\begin{document}

\twocolumn[
\icmltitle{Deconfounded Representation Similarity for Comparison of Neural Networks}


\icmlsetsymbol{equal}{*}

\begin{icmlauthorlist}
\icmlauthor{Tianyu Cui}{aalto}
\icmlauthor{Yogesh Kumar}{aalto}
\icmlauthor{Pekka Marttinen}{equal,aalto}
\icmlauthor{Samuel Kaski}{equal,aalto,manch}
\end{icmlauthorlist}

\icmlaffiliation{aalto}{Department of Computer Science, Aalto University, Espoo, Finland}
\icmlaffiliation{manch}{Department of Computer Science, University of Manchester, Manchester, UK}

\icmlcorrespondingauthor{Tianyu Cui}{tianyu.cui@aalto.fi}

\icmlkeywords{Neural networks, representation similarity}

\vskip 0.3in
]



\printAffiliationsAndNotice{\icmlEqualContribution} 

\begin{abstract}

Similarity metrics such as representational similarity analysis (RSA) and centered kernel alignment (CKA) have been used to compare  layer-wise representations between neural networks. However, these metrics are confounded by the population structure of data items in the input space, leading to spuriously high similarity for even completely random neural networks and inconsistent domain relations in transfer learning. We introduce a simple and generally applicable fix to adjust for the confounder with covariate adjustment regression, which retains the intuitive invariance properties of the original similarity measures. We show that deconfounding the similarity metrics increases the resolution of detecting semantically similar neural networks. Moreover, in real-world applications, deconfounding improves the consistency of representation similarities with domain similarities in transfer learning, and increases correlation with out-of-distribution accuracy.

\end{abstract}

\section{Introduction}
\label{sec: intro}
Deep neural networks (NNs) have achieved state-of-the-art performance on a wide range of machine learning tasks by automatically learning feature representations from data \cite{krizhevsky2012imagenet,wang2018glue,irvin2019chexpert,rajpurkar2016squad,lin2014microsoft}. However, these networks do not offer interpretable predictions on most applications and are seen as ``black boxes''. It is thus crucial to understand the intricacies of neural networks before they are deployed on critical applications. Previous work has made progress in understanding how a single neural network makes decisions with axiomatic attribution methods \cite{sundararajan2017axiomatic,lundberg2017unified} and understanding how multiple neural networks relate to each other with representation similarity measures \cite{mehrer2020individual}. 

Several similarity measures between representations have been proposed with different principles, including linear regression \cite{romero2014fitnets}, canonical correlation analysis (CCA; \citealt{raghu2017svcca, morcos2018insights}), statistical shape analysis \cite{williams2021generalized}, and functional behaviors on down-stream tasks \cite{alain2016understanding,feng2020transferred,ding2021grounding}. Another main-stream approach is based on representational similarity analysis, RSA \cite{edelman1998representation,kriegeskorte2008representational,mehrer2020individual,shahbazi2021using}, which computes the similarity between 
(dis)similarity matrices of two neural network representations of the same dataset, such as centered kernel alignment (CKA, \citealt{kornblith2019similarity}).

Despite the wide usage of RSA and CKA in understanding biological \cite{haxby2001distributed} and artificial neural networks \cite{nguyen2020wide}, we find that the inter-example (dis)similarity matrices in the representation space of different NNs are highly correlated with a shared factor (i.e., a confounder): the (dis)similarity structure of the data items in the input space, especially for shallow layers. This confounding issue limits the ability of CKA to reveal similarity of models on the functional level. This leads to spuriously high CKAs even between two random neural networks, and counter-intuitive conclusions when comparing CKAs on sets of models trained on different domains, such as in the transfer learning setting \cite{neyshabur2020being,kornblith2021better}

In this paper, we propose a simple approach to adjust the representation similarity by regressing out the confounder, the inter-example (dis)similarity matrix in the input space, from the (dis)similarity matrices of two representations. This is inspired by the covariate adjusted correlation analysis widely studied in Biostatistics \cite{wu2018covariate,liu2018covariate}. This approach can be applied to any similarity measure built on the RSA framework. Moreover, we study the invariance properties of the deconfounded representation similarity and demonstrate its benefits on public image and natural language datasets with various neural network architectures. 

Overall, our contributions are:
\begin{itemize}
\item We study the confounding effect of the input inter-example similarity on the representation similarity between two neural networks, and propose a simple and generally applicable deconfounding fix. We discuss the invariance properties of the deconfounded similarities.
\item We verify that deconfounded similarities can detect semantically similar neural networks from random neural networks, and small model changes across domain tasks where previous similarity measures fail.
\item We show that deconfounded similarities are more consistent with domain similarities in transfer learning, compared with existing methods.
\item We demonstrate that deconfounded similarities on in-distribution datasets are more correlated with out-of-distribution accuracy than the corresponding original similarities \cite{ding2021grounding}.
\end{itemize}

\section{Preliminaries}
\label{sec: probsetup}
\begin{figure*}[t]
    \centering
    \includegraphics[width=1\linewidth]{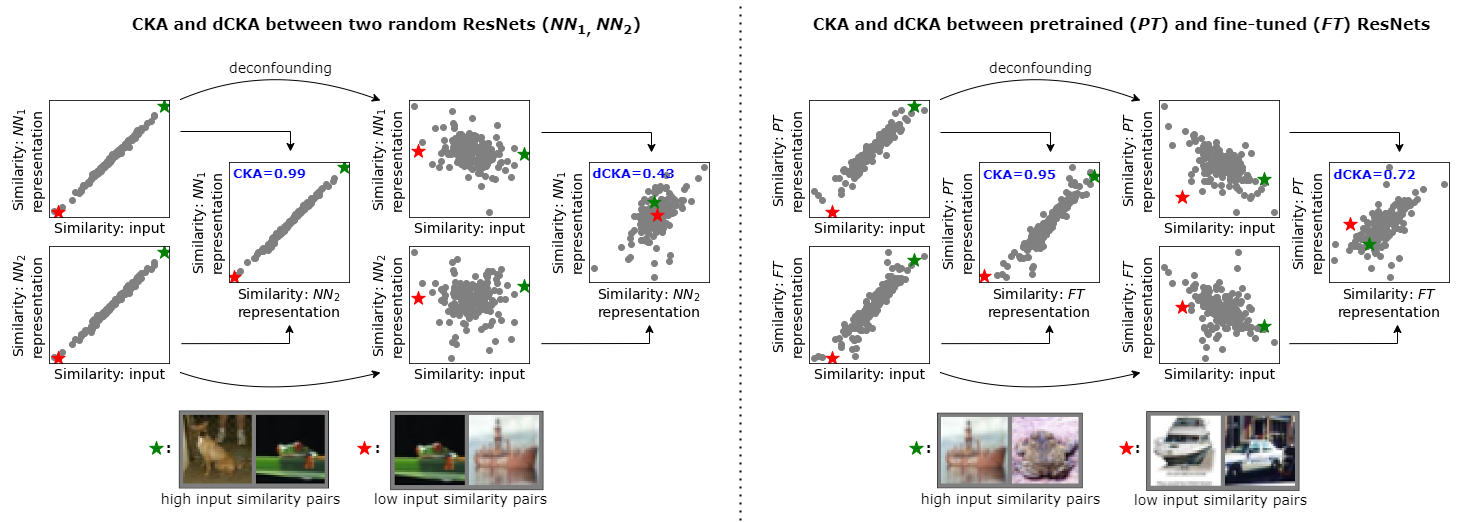}
    \caption{\textbf{Demonstration of the confounder in CKA.} CKA calculates the similarity between inter-example similarities for two representations, which are confounded by the inter-example similarities in the input space, such that input pairs with high (\greenstar) and low (\redstar) input similarities also have high and low representation similarities on both random NNs (\textit{Left}) and trained NNs (\textit{Right}) representations. Moreover,  the confounder leads to the counterintuitive conclusion that CKA on random NNs is higher than pretrained and fine-tuned NNs on similar domains (0.99 vs. 0.95). This is resolved by deconfounding (0.43 vs.0.72).}
    \label{fig:dCKA demo}
    \vskip -0.2in
\end{figure*}
\subsection{Notation and prior work}
 Let $X\in\mathbb{R}^{n\times p}$ denote the input dataset with $n$ datapoints and $p$ features, and $X^{m_1}_{f_{1}}\in\mathbb{R}^{n\times p_1}$ and $X^{m_2}_{{f_2}}\in\mathbb{R}^{n\times p_2}$ denote the $m_1$th and $m_2$th layer representations of two neural networks of interest, $f_{1}(X)$ and $f_{2}(X)$, with $p_1$ and $p_2$ nodes respectively. We center and normalize the representation matrices by first removing the mean of each feature (i.e., each column), and then dividing by the Frobenius norm. 
 
 A standard approach for comparing representations of two neural networks is to compare the similarity structures in each network representation. This can be done by first computing the similarity between every pair of examples in $X^{m_1}_{{f_1}}$ and $X^{m_2}_{{f_2}}$ with a similarity measure $k(\cdot,\cdot)$:
\begin{equation}
\label{eq: similarity structure}
\begin{split}
&K^{m_1}_{{f_1}}=k(X^{m_1}_{{f_1}}, X^{m_1}_{{f_1}})\\
&K^{m_2}_{{f_2}}=k(X^{m_2}_{{f_2}}, X^{m_2}_{{f_2}}).
\end{split}
\end{equation}
Here $K^{m_1}_{{f_1}}, K^{m_2}_{{f_2}}\in\mathbb{R}^{n\times n}$ are called representational similarity matrices (RSMs)\footnote{Note that $k(\cdot,\cdot)$ can also be dissimilarity measure, but we call $K^{m}_{f}$ a similarity matrix for simplicity.}. Second, another similarity measure $s(\cdot,\cdot)$ is applied to compare these two similarity structures, $K^{m_1}_{{f_1}}$ and $K^{m_2}_{{f_2}}$:
\begin{equation}
\label{eq: similarity between similarity}
s^{m_1,m_2}_{{f_1},{f_2}} = s(K^{m_1}_{{f_1}},K^{m_2}_{{f_2}}).
\end{equation}
This gives the similarity between the two representations.

The existing approaches vary by using different similarity measures for both levels of comparison. CKA \cite{kornblith2019similarity} employs a kernel function for the first level similarity $k(\cdot,\cdot)$, and an Hilbert-Schmidt Independence Criterion (HSIC) estimator for the second, $s(\cdot,\cdot)$. On the other hand, RSA-based methods use Euclidean distance to measure the inter-example \textbf{dis}similarity structure, and apply Pearson's correlation \cite{mehrer2020individual} or Spearman's rank correlation \cite{shahbazi2021using} to quantify the similarity between two dissimilarity structures.

Although other approaches, e.g., linear regression \cite{yamins2014performance} and CCA \cite{raghu2017svcca}, have been proposed to compare neural network representations, we focus on CKA and RSA in this paper due to their wide usage in understanding the properties of neural networks, such as transfer learning \cite{neyshabur2020being,kornblith2021better,raghu2021vision}.

\subsection{Illustration of the confounding in representation similarity}

As shown in Eq.\ref{eq: similarity between similarity}, the similarity between two RSMs, $K^{m_1}_{{f_1}}$ and $K^{m_2}_{{f_2}}$, defines the similarity between two neural networks. However, both these similarity structures are affected by the same factor (i.e., a confounder): the similarity structure of input dataset $X$, which can cause spuriously high similarity. Intuitively, (dis)similar data points (green stars and red stars in Figure \ref{fig:dCKA demo}) in the input space are likely to be (dis)similar in the representation space of the first few layers. Thus the representation similarity structure of different neural networks would be similar even for random neural networks that have totally different functional behaviors. This is undesirable since the goal of calculating the similarity measures for neural networks is to quantify how similar the networks are, and ideally this should not be affected by the specifics of the dataset at hand.

Figure \ref{fig:dCKA demo} illustrates this problem using the CKA similarity measure as an example, and compares that with the decounfounded dCKA (defined in the next section) on the first-layer of ResNets \cite{he2016deep} with 20 random samples from CIFAR-10 \cite{krizhevsky2009learning} test set. We consider two pairs of ResNets: 1. two random ResNets generated by adding different Gaussian noise, $\mathcal{N}(0,1)$, to each parameter of the pretrained ResNet-18\footnote{\url{https://pytorch.org/vision/stable/models.html}} on ImageNet \cite{deng2009imagenet}; 2. the pretrained (PT) ResNet-18 and a fine-tuned (FT) ResNet-18 on CIFAR-10. We notice that CKA on random NNs is almost 1, and counterintuitively it is even higher than the CKA between PT and FT ResNets on a similar domain (0.99 vs. 0.95), although we would expect the PT and FT networks to learn similar low-level features and hence be more similar than random networks. This happens because the similarities between samples in the input space confound their similarities in the representation space. After adjusting for the confounder with dCKA, the similarity between the two random ResNets is much smaller than the similarity of the PT and FT networks (0.43 vs. 0.72).

\section{Methods}
\subsection{Deconfounding representation similarity}

We propose a simple approach to adjust the spurious similarity caused by the confounder by regressing out the input similarity structure from the representation similarity structure \cite{csenturk2005covariate}. That is:
\begin{equation}
\begin{split}
&dK^{m_1}_{{f_1}}=K^{m_1}_{{f_1}}-\hat{\alpha}^{m_1}_{f_1}K^{0};\\
&dK^{m_2}_{{f_2}}=K^{m_2}_{{f_2}}-\hat{\alpha}^{m_2}_{f_2}K^{0},
\label{eq: deconfound similarity structure}
\end{split}
\end{equation}
where the $\hat{\alpha}^{m_1}_{f_1}$ and $\hat{\alpha}^{m_2}_{f_2}$ are the regression coefficients that minimize the Frobenius norm of $dK^{m_1}_{{f_1}}$ and $dK^{m_2}_{{f_2}}$ respectively. Furthermore, the letter $d$ is front of a similarity matrix, e.g. as in $dK^{m_1}_{{f_1}}$, denotes the deconfounded version of $K^{m_1}_{{f_1}}$, and similarly the letter $d$ is applied throughout the text to denote all defounded quantities. To do the deconfounding, we assume that the input similarity structure $K^{0}$ has a linear and additive effect on $K^{m}_{{f}}$, i.e.,
\begin{equation}
\text{vec}(K^{m}_{{f}})=(\alpha^{m}_{f})^T\text{vec}(K^{0})+\boldsymbol{\epsilon}^{m}_{f},
\label{eq: linear regression}
\end{equation}
where noise $\boldsymbol{\epsilon}^{m}_{f}$ is assumed to be independent from the confounder with $\boldsymbol{\hat{\epsilon}}^{m}_{f}=\text{vec}(dK^{m}_{{f}})$ and
\begin{equation}
\hat{\alpha}^{m}_{f}=(\text{vec}(K^{0})^{T}\text{vec}(K^{0}))^{-1}\text{vec}(K^{0})^{T}\text{vec}(K^{m}_{{f}}).
\label{eq: linear regression coef}
\end{equation}
After the deconfounded similarity structures are obtained with Eq.\ref{eq: deconfound similarity structure}, we use the same similarity measure to calculate the deconfounded representation similarity:
\begin{equation}
\label{eq: deconfound similarity}
ds^{m_1,m_2}_{{f_1},{f_2}} = s(dK^{m_1}_{{f_1}},dK^{m_2}_{{f_2}}).
\end{equation}

Note that $dK^{m}_{{f}}$ is not always positive semi-definite, even when $K^{m}_{{f}}$ is positive semi-definite (PSD). For a similarity measure $s(\cdot,\cdot)$ that takes two kernel matrices as input, such as the CKA, we transform $dK^{m}_{{f}}$ into a positive semi-definite matrix by removing all the negative eigenvalues according to \citep{chan1997algorithm}. Specifically, we have the eigenvalue decomposition of $dK^{m}_{{f}}$, such that
\begin{equation}
\begin{split}
&dK^{m}_{{f}} = Q\Lambda Q^{T}=Q(\Lambda_{+}-\Lambda_{-})Q^{T},\\
\Lambda_{\pm}&=\text{diag}\{\max(0,\pm\lambda_1),\ldots,\max(0,\pm\lambda_n)\},
\end{split}
\end{equation}
where $\lambda_i$ is the $i$th eigenvalue of $dK^{m}_{{f}}$. We approximate $dK^{m}_{{f}}$ with a PSD matrix $\Tilde{dK^{m}_{{f}}}$:
\begin{equation}
dK^{m}_{{f}}\approx \Tilde{dK^{m}_{{f}}} = \rho^2Q\Lambda_{+}Q^{T};\;\;\rho=\frac{|\text{tr}(\Lambda)|}{|\text{tr}(\Lambda_{+})|}.
\label{eq:psd correction}
\end{equation}

\subsection{Examples of deconfounded similarity indices}
\textbf{Deconfounded CKA.} In CKA \cite{kornblith2019similarity}, the similarity structure in the feature space is represented with a valid kernel $l(\cdot,\cdot)$, i.e., $K^{m_1}_{{f_1}} = l(X_{f_1}^{m_1},X_{f_1}^{m_1})$ and $K^{m_2}_{{f_2}} = l(X_{f_2}^{m_2},X_{f_2}^{m_2})$, such as the linear or RBF kernel. Then an empirical estimator of HSIC \cite{gretton2005measuring} is used to align two kernels: 
\begin{equation}
\text{HSIC}^{m_1,m_2}_{f_1,f_2}=\frac{1}{(n-1)^2}\text{tr}(K^{m_1}_{{f_1}}HK^{m_2}_{{f_2}}H),
\label{eq:hsic}
\end{equation}
where $H$ is the centering matrix. CKA is given by the normalized HSIC such that
\begin{equation}
\text{CKA}(K^{m_1}_{{f_1}},K^{m_2}_{{f_2}})=\frac{\text{HSIC}^{m_1,m_2}_{f_1,f_2}}{\sqrt{\text{HSIC}^{m_1,m_1}_{f_1,f_1}\text{HSIC}^{m_2,m_2}_{f_2,f_2}}}.
\label{eq:cka}
\end{equation}

To deconfound the representation similarity matrices $K^{m_1}_{{f_1}}$ and $K^{m_2}_{{f_2}}$, we apply the same kernel to measure the inter-example similarity in the input space $K^0=l(X,X)$, and adjust its confounding effect with Eq.\ref{eq: deconfound similarity structure}. However, matrices $dK^{m_1}_{{f_1}}$ and $dK^{m_2}_{{f_2}}$, obtained by regressing out one kernel matrix from another kernel, are no longer kernels, and they are not applicable for computing HSIC. Fortunately, with Eq.\ref{eq:psd correction}, we can approximate the $dK^{m_1}_{{f_1}}$ and $dK^{m_1}_{{f_1}}$ with two valid kernels $\Tilde{dK^{m_1}_{{f_1}}}$ and $\Tilde{dK^{m_2}_{{f_2}}}$, which are then used to construct the deconfounded CKA (dCKA):
\begin{equation}
\text{dCKA}(K^{m_1}_{{f_1}},K^{m_2}_{{f_2}})=\text{CKA}(\Tilde{dK^{m_1}_{{f_1}}},\Tilde{dK^{m_2}_{{f_2}}}).
\label{eq:dcka}
\end{equation}
We use a linear kernel here because \citet{kornblith2019similarity} report similar results with using a RBF kernel.

\textbf{Deconfounded RSA.} Different from CKA, the similarity structure in RSA \cite{mehrer2020individual} is measured by the pairwise Euclidean distance between examples in the feature space. Specifically, each element of $K^{m_1}_{f_1}$ and $K^{m_2}_{f_2}$ is obtained by $K^{m_1}_{f_1,ij}=\lVert\boldsymbol{x}^{m_1}_{f_1,i} - \boldsymbol{x}^{m_1}_{f_1,j}\rVert^2$ and $K^{m_2}_{f_2,ij}=\lVert\boldsymbol{x}^{m_2}_{f_2,i} - \boldsymbol{x}^{m_2}_{f_2,j}\rVert^2$, where $\boldsymbol{x}^{m_1}_{f_1,i}$ is the $m_1$-layer representation of the $i$th example in neural network $f_1$. Thus, the input similarity structure $K^{0}$ is measured with the pairwise Euclidean distance in the input space. After $K^{0}$ is adjusted with Eq.\ref{eq: deconfound similarity structure}, we apply Spearman's $\rho$ correlation to measure the similarity between the upper triangular part of $dK^{m_1}_{f_1}$ and $dK^{m_2}_{f_2}$, i.e., $\text{triu}(dK^{m_1}_{{f_1}})$ and $\text{triu}(dK^{m_2}_{{f_2}})$, that is
\begin{equation}
\text{dRSA}(K^{m_1}_{{f_1}},K^{m_2}_{{f_2}})=\rho(\text{triu}(dK^{m_1}_{{f_1}}),\text{triu}(dK^{m_2}_{{f_2}})).
\label{eq:drsa}
\end{equation}
Note that rank correlation does not require two similarity matrices to be positive semi-definite. Therefore, we skip the steps of constructing the PSD approximation.
\subsection{Other details}
\begin{figure}[t]
\begin{center}
\includegraphics[width=\columnwidth]{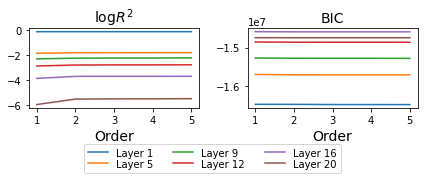}
\vskip -0.1in
\caption{\textbf{Log $R^2$ and BIC of regression models with different orders of input similarity}. We empirically observe that correcting for the confounder with a linear model is sufficient, especially for shallow layers (demonstrated with ResNet-18 on CIFAR-10).}
\label{fig: optimal order}
\end{center}
\vskip -0.2in
\end{figure}
\textbf{Is linear model sufficient?} In Eq.\ref{eq: linear regression}, we assume that the representation similarity structures \textbf{linearly} depend on the input similarity structure. To validate if the linear assumption is sufficient, we check if adding higher-order polynomial terms of input similarity to the regression model (Eq.\ref{eq: linear regression}) can help explain the representation similarity structure. 

We show the effect of adding higher-order polynomial terms on a pretrained ResNet-18 (contains 20 layers in total) with CIFAR-10 inputs in Figure \ref{fig: optimal order}. We observe that neither the $R^2$ nor the Bayesian information criterion (BIC) \cite{murphy2012machine}, approximating the model evidence, changes much when adding higher-order terms. Although $R^2$ can be marginally improved by increasing the order in the deeper layers (e.g., layer 20), we only consider the linear model in this paper for simplicity.

\textbf{Does the independent noise assumption hold?} In Eq.\ref{eq: linear regression}, the regression targets are similarities between every pair of examples. Thus, noise $\epsilon_{f, ij}^{m}$ of example-pair $i,j$ might be correlated with $\epsilon_{f, ik}^{m}$ of example-pair $i,k$ because they both are associated with the same example $i$. However, the Durbin-Watson tests \cite{durbin1992testing} show that the independent noise assumption still holds (Appendix \ref{app: DW test}).
\subsection{Theoretical properties}
In this section, we study the invariance properties of the deconfounded representation similarity. For similarity measures that we studied, i.e., CKA and RSA, the corresponding deconfounded similarity indices have the same invariance properties, such as invariance to orthogonal transformation and isotropic scaling,  as the original similarity measures. 

\begin{proposition}
\label{proposition: orthogonal transformation invariance}
Deconfounded CKA and deconfounded RSA are invariant to orthogonal transformation, if the (dis)similarity measure $k(\cdot,\cdot)$  that compare inter-examples are orthogonal invariant.
\end{proposition}
\begin{proposition}
\label{proposition: isotropic scaling invariance}
Deconfounded CKA with a linear kernel and deconfounded RSA are invariant to isotropic scaling.
\end{proposition}
Intuitively, as long as $k(\cdot,\cdot)$ is invariant to orthogonal transformation, e.g., linear kernels and Euclidean distance, the deconfounded representation similarity matrix $dK^{m}_{f}$ in Eq.\ref{eq: deconfound similarity structure} is also invariant to orthogonal transformation, because it is defined in terms of the kernel $k$. Thus all operations on $dK^{m}_{f}$ are invariant to orthogonal transformation. Moreover, if one representation is scaled by a scalar,  $dK^{m}_{f}$ and $\Tilde{dK^{m}_{f}}$ will be scaled by the same scalar, whose effects will be finally eliminated in the normalization step in CKA (Eq.\ref{eq:cka}) and the rank correlation step in RSA (Eq.\ref{eq:drsa}). 

\section{Experiments}
\subsection{Consistency of layer-wise similarities for NNs trained with different initialization}
\begin{figure}[t]
\begin{center}
\includegraphics[width=1\columnwidth]{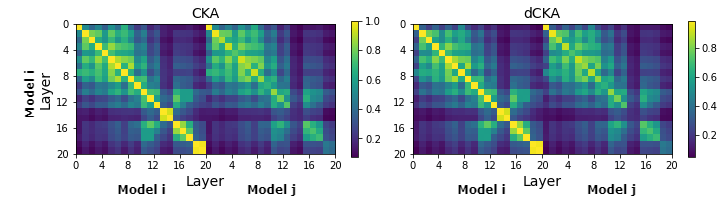}
\caption{\textbf{CKA and dCKA between layers}. \textit{Left} panel: layers are from the same NN. \textit{Right} panel: layers are from NNs trained with different initializations. Like CKA, dCKA can reveal consistent relationships between layers of NNs trained with different initializations.}
\label{fig: compare cka dcka}
\end{center}
\vskip -0.2in
\end{figure}
One advantage of original CKA is that it can reveal consistent relationships between layers of neural networks trained with different initializations, while other representation similarity measures, such as PWCCA and Procrustes, can not \cite{kornblith2019similarity,ding2021grounding}. Here we show that dCKA has a similar behavior as CKA when studying the similarity between representations of neural networks with different initializations.

As \citet{ding2021grounding}, we first take 5 ResNet-18 models trained on CIFAR-10 dataset with different initializations. For one model (model $i$), we compute the similarity between every pair of layers of the model. We then choose another model with a different initialization (model $j$), and compute the similarity between every layer of model $i$ and every layer of model $j$. We average the results for five different models.

We show the results in Figure \ref{fig: compare cka dcka}, where we observe that for both CKA and dCKA and for each layer of the model $i$, the most similar layer in model $j$ is the same corresponding layer. Hence, similarly to CKA, dCKA can identify consistent relationships of layers between different networks.

\subsection{Ability to detect pairs of similar networks from pairs of random networks}
\label{sec: detect similar networks}
We propose a simple check to see if similarity measures can identify similar neural network representations from random neural network representations. We first construct 50 random ResNets, and compute the similarities between every pair of random neural networks at each model block (containing 2-3 convolutional layers) on CIFAR-10 test set, which gives the null distribution. Given a pretrained ResNet-18 on ImageNet, we consider two ways of generating random neural networks: 1. permute the weight matrix of each layer; 2. add large Gaussian noise, $\mathcal{N}(0,10)$, to each weight per layer. We then train 50 ResNet-18 networks with different random initializations from scratch on CIFAR-10 (shown in Appendix \ref{sec: training details}), and compute the similarity between each CIFAR Resnet with the ImageNet ResNet (i.e., the pretrain ResNet) on the same on CIFAR-10 test set for each block, and this gives the alternative distribution of similarities. 

Intuitively, the alternative distribution of similarities should be significantly larger than the corresponding null distribution especially for shallow blocks, because models trained on ImageNet and CIFAR should learn similar low-level representations \cite{goodfellow2016deep}. We compute the proportion (shown in Table \ref{tab: null-distribution}) of 50 ImageNet-CIFAR ResNets pairs whose similarities are significantly larger than the null distributio, i.e., larger than the upper bound of its 95\% CI.

\begin{table}[t]
\caption{\textbf{Proportion of ImageNet-CIFAR ResNets pairs identified from random ResNets}. For each block of ResNets, ImageNet-CIFAR pairs are identified if the similarities are significantly larger than the corresponding null distribution generated by adding noise and permutation (the first and second numbers) to the ImageNet ResNet. We observe deconfounded similarities can improve the identification of semantically similar NNs from random NNs. }
\vskip 0.1in
\label{tab: null-distribution}
\centering
\footnotesize
\begin{tabular}{ccccc}
\toprule
Block   & CKA       & dCKA     & RSA       & dRSA      \\ \hline
 1 & 1.0, 1.0       & 1.0, 1.0      & 0.18, 0.88 & 1.0, 1.0       \\
 2 & 0.94, 1.0    & 1.0, 1.0      & 0.0, 0.28    & 1.0, 1.0       \\
 3 & 0.0, 1.0       & 1.0, 1.0      & 0.0, 0       & 0.0, 0.08    \\
 4 & 0.0, 0.28    & 1.0, 1.0      & 0.0, 0.0       & 0.0, 0.0       \\
 5 & 0.0, 0.0       & 0.0, 0.04   & 0.0, 0.0       & 0.0, 0.0       \\
 6 & 0.0, 0.0       & 0.0, 0.02   & 0.0, 0.0       & 0.0, 0.0       \\
 7 & 0.0, 0.0       & 0.0, 0.0      & 0.0, 0.0       & 0.0, 0.0       \\
 8 & 0.0, 0.0       & 0.0, 0.0      & 0.0, 0.0       & 0.0, 0.0       \\ \hline
Average & 0.24, 0.41 & \textbf{0.5, 0.51} & 0.02, 0.15 & \textbf{0.25, 0.26} \\ \bottomrule
\end{tabular}
\end{table}

\begin{figure}[t]
\begin{center}
\includegraphics[width=\columnwidth]{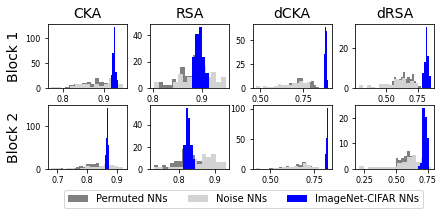}
\caption{\textbf{Histogram of each similarity measure for the first two blocks.} Compared with CKA and RSA, dCKA and dRSA improve separating ImageNet-CIFAR pairs from random networks, even the proportion of identified similar network pairs remains unchanged.}
\label{fig: null_distribution hist}
\end{center}
\vskip -0.2in
\end{figure}

In Table \ref{tab: null-distribution}, we observe that in shallow blocks the deconfounded similarity can detect a larger proportion of the ImageNet-CIFAR model pairs from the random neural network pairs than the other dimilarity measures. For example, in the fourth block, dCKA can still detect all correct pairs whereas the original CKA detects only 0 or 0.28 of pairs from the two types of random pairs, correspondingly. In Figure \ref{fig: null_distribution hist}, we visualize histograms of each similarity measure under the two types of null distributions and the alternative distribution for the first two blocks. We observe that although CKA and dCKA or RSA and dRSA can all identify similar proportions in block 1 and 2 with certain null distributions, the difference between the correct and random pairs is more significant with the deconfounded similarities. Moreover, in deep layers, e.g., after block 7, no method can identify ImageNet-CIFAR pairs anymore. We hypothesise that ImageNet and CIFAR-10 contain different classes of images, thus their high-level representations are significantly different. We discuss this more in Appendix \ref{app: detect similar NNs}.

\subsection{Consistency of NN similarities across domains}
\label{sec: ood consistency test}
Ideally, similarity of neural networks would not depend on the domain in which the networks are applied. Here, we study this by constructing a set of 6 neural networks, $\{f_i|i\in\{1,2,3,4,5,6\}\}$, by adding independent Gaussian noise $i\times\mathcal{N}(0,0.1)$ to each parameter of the pretrained ResNet-18, $f^{*}$. Hence, the similarity $s(f_i, f^{*})$ should be higher than $s(f_{i+1}, f^{*})$ regardless of the input domain in which the the networks are applied to calculate representations. Thus, we calculate the similarity $s(f_i, f^{*})$ for every $f_i$ on each domain of the corrupted CIFAR-10-C dataset \cite{hendrycks2019benchmarking} that contains 19 domains with different types of corruptions to the original CIFAR-10. We compute the average similarity $\mu_{f_i,f^*}$ across the 19 domains and its standard error $\sigma_{f_i,f^*}$. We say $f_i$ is significantly more similar to $f^{*}$ than to $f_{i+1}$ across domains, if
\begin{equation}
\mu_{f_i,f^*}-1.96\sigma_{f_i,f^*} > \mu_{f_{i+1},f^*}+1.96\sigma_{f_{i+1},f^*}.
\label{eq:across domain test}
\end{equation}
We repeat the above experiments 20 times with different random seeds, i.e., generate 20 different sets of neural networks, to measure the proportion of cases where $f_i$ is significantly more similar to $f^{*}$ than $f_{i+1}$ for each block, as well as the confidence interval of the proportion.

\begin{figure}[t]
\begin{center}
\includegraphics[width=\columnwidth]{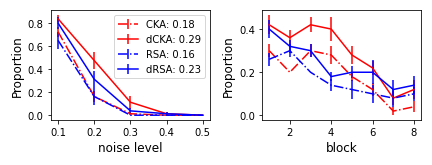}
\caption{\textbf{Proportion of identified similar NNs across different domains.} The proportion is summarized for different noise levels on \textit{Left} and different NN blocks on \textit{Right}. We observe that the deconfounded similarity can identify more similar models compared with the corresponding original similarity. Moreover, NNs are harder to identify for higher noise level and deeper layers.}
\label{fig: noisy NNs across domain}
\end{center}
\vskip -0.2in
\end{figure}

\begin{figure}[t]
\begin{center}
\includegraphics[width=\columnwidth]{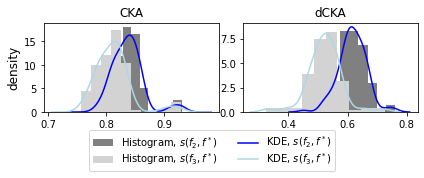}
\caption{\textbf{Histogram and kernel density estimation (KDE) of CKA and dCKA across 19 domains on the first-block representations}. We observe that dCKA can separate $s(f_2, f^*)$ from $s(f_3, f^*)$ better than CKA.}
\label{fig: noisy NNs across domain hist}
\end{center}
\vskip -0.2in
\end{figure}

In Figure \ref{fig: noisy NNs across domain} \textit{Left}, we show the proportion of identified NNs averaged over all blocks for each noise level. We observe that the deconfounded similarity improves the proportion of identified NNs compared with the corresponding original similarity for all noise levels. The averaged proportion increases $59.7\%$ for CKA (from 0.18 to 0.29) and $43.1\%$ for RSA (from 0.16 to 0.23) after deconfounding. We also observe that deconfounding can improve CKA/RSA on different inputs from the \textit{same} domain, but the improvement is relatively smaller ($23\%$ for CKA, from 0.65 to 0.8, and $7\%$ for RSA, from 0.75 to 0.8), shown in Appendix \ref{app: consistency in-domain}. Moreover, the proportion decreases as the noise level increases for all similarity measures, because for large noise level (large $i$), both $f_{i+1}$ and $f_i$ are far $f^{*}$. In Figure \ref{fig: noisy NNs across domain} \textit{Right}, we show the the proportion of consistently identified similarities for each block where the results are averaged over different noise levels. In general, we expect to identify fewer similar NNs with deeper layer representations, because the Gaussian noise is added to each parameter and deeper representations consequently accumulate more noise than shallow layers. 

We visualize the histogram of CKA and dCKA between $f_2$ and $f^{*}$ and between $f_3$ and $f^{*}$ on the first-block representations of inputs from 19 domains in Figure \ref{fig: noisy NNs across domain hist}, where we can clearly observe that $f_2$ and $f_3$ are more separatable in terms of dCKA than CKA.

\subsection{Transfer learning: domain similarity vs. PT and FT similarity }
\label{sec:domain-shift}

\citet{ding2021grounding} argued that the similarity metric must be sensitive to changes that affect the functionality of the networks we compare. We extend this to transfer learning under domain shift, where we expect the similarity metric to be sensitive to domain changes. Consider two models with the same initialization from pretrained weights which are finetuned on data from different domains. We then expect the similarity of the layer representations between the finetuned model and the initial pretrained model to be different for each domain. Further, this representation similarity between the finetuned and pretrained models should depend on the similarity between the source and target domains. 

To study this in detail for dCKA and CKA, we choose datasets from two modalities -- image and text -- that display such covariate domain shift. For text, we use the Multi-lingual STS-B dataset \cite{huggingface:dataset:stsb_multi_mt} and choose English, Spanish, Polish and Russian languages as the target domains. For images, we use the Real, Clipart, Sketch and Quickdraw as target domains from the DomainNet dataset \cite{peng2019moment}. Next, we finetune separate models for each domain from both modalities. For text, we initialize a pretrained XLM-RoBERTa \cite{conneau2020unsupervised} model and for images, we pretrain a ResNet-50 model on Imagenet \citep{deng2009imagenet}. 

In order to quantify the similarity between two domains, we build a cross-domain classifier  \cite{rabanser2018failing} and create a dataset using equal number of samples from each domain (we use 80\%  of the data for training and 20\% for validating the generalization). We then train a weak discriminator model to predict the appropriate true domain for each sample. For the discriminator, we use EfficientNet-B0 \citep{tan2019efficientnet} for images and Distil-RoBERTa \citep{Sanh2019DistilBERTAD} for text. The discriminator is essentially a binary classifier and thus the test binary cross entropy (BCE) loss would be an indicator of the similarity between the two domains (higher meaning more similar). 

\begin{figure*}[ht]
  \centering
  \includegraphics[width=0.95\linewidth]{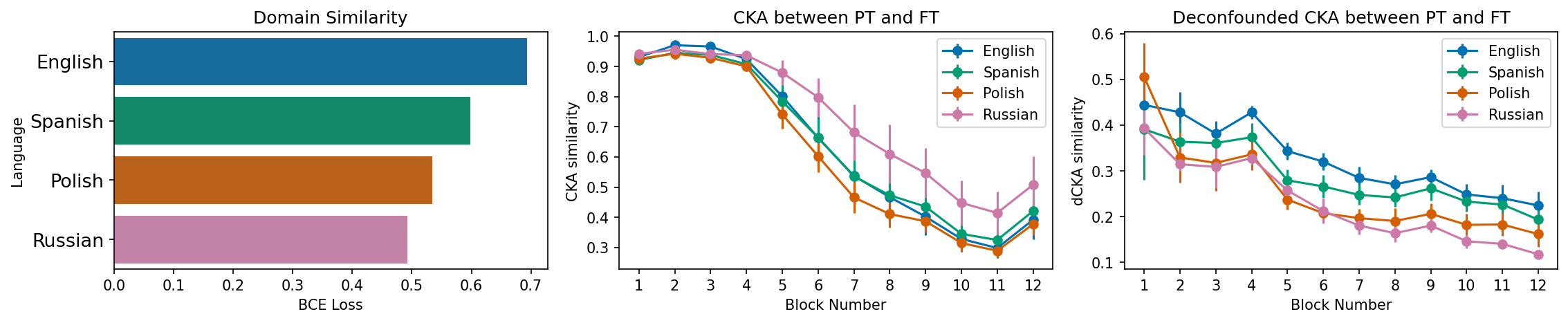}
  \caption[]{\textbf{dCKA adjusts for transfer learning under domain shift.} \emph{Left} shows the ground truth domain similarity (between English and other languages) as measured by test binary cross entropy (BCE) loss of the cross-domain classifier. We plot the CKA (\emph{Center}) and dCKA (\emph{Right}) between the pretrained XLM-RoBERTa model and models finetuned for different languages on the STS-B task. We observe that the representation similarity measured by dCKA is better correlated with the domain similarity.}
  \label{fig:domain-transfer}
  \vskip -0.2in
\end{figure*}

Figure \ref{fig:domain-transfer} \emph{Left} shows the results for the test BCE loss when the cross-domain classifier was trained to discriminate between the domains on STS-B -- english-english, english-spanish, english-polish and english-russian. The layer-wise CKA and dCKA between the pretrained and finetuned XLM-RoBERTa model is shown Figure \ref{fig:domain-transfer} (\emph{Center} and \emph{Right}), respectively. The models finetuned on highly dissimilar domains (e.g., english-russian) are expected to have lower layer-wise similarity with the pretrained model. We see that dCKA captures this relationship better than CKA. More concretely, we measure the Spearman's $\rho$ and Kendall's $\tau$ correlation between the cross-domain classifier BCE loss and the representational similairty measured by CKA and dCKA between the initial pretrained model and the finetuned model. Table \ref{tab:domain-transfer-results} summarizes the results for finetuned models from 10 random restarts. We can see that the dCKA is more correlated with the domain similarity as compared to CKA on both modalities.

\begin{table}[]
\centering
\footnotesize
\caption{\textbf{Rank correlation between the domain similarity and the representation similarity between pretrained and finetuned models}. We see that compared to CKA, the dCKA has higher correlation with domain similarity in terms of Spearman's $\rho$ and Kendall's $\tau$.}
\label{tab:domain-transfer-results}
\vskip 0.1in
\begin{tabular}{ccccc}
\toprule
\multirow{2}{*}{Modality} & \multicolumn{2}{c}{CKA}   & \multicolumn{2}{c}{dCKA} \\ 
\cmidrule(l){2-5} 
\multicolumn{1}{c}{}    & $\rho$ &  $\tau$  & $\rho$    & $\tau$ \\ 
\midrule \medskip
DomainNet               & 0.804  & 0.628    & \textbf{0.895}     & \textbf{0.780}  \\ 

Multi STS-B             & -0.358 & -0.248 & \textbf{0.924} & \textbf{0.809}  \\
\bottomrule
\end{tabular}
\end{table}

\subsection{Out-of-distribution accuracy}
Here we evaluate the sensitiveness of deconfounded similarities to changes that affect the functionality of neural networks on out-of-distribution (OOD) data, and we would expect similar NNs to have similar OOD accuracy. We follow a similar experiment setup as previous work \cite{ding2021grounding}: 1. We train 50 ResNet-18, $f_i$, with different random initialization on CIFAR-10; 2. Evaluate the OOD accuracy of each model on CIFAR-10-C \cite{hendrycks2019benchmarking}, $\text{acc}(f_i)$, and select the most accurate ResNet as the reference model, $f^{*}$; 3. Compute the similarity between each $f_i$ and $f^{*}$, $s(f_i,f^{*})$, of each block on \textbf{CIFAR-10} test set, and compute the accuracy difference on \textbf{CIFAR-10-C} test set $|\text{acc}(f_i)-\text{acc}(f^{*})|$; 4. Measure the rank correlation, e.g., Kendall's $\tau$ and Spearman's $\rho$, between $1-s(f_i,f^{*})$ and $|\text{acc}(f_i)-\text{acc}(f^{*})|$ for each block. A good similarity should have a high rank correlation, meaning that the similarity in the input space correlates with OOD accuracy.

\begin{table}[t]
\caption{\textbf{Rank correlation with standard error (in parentheses) between representation similarity and prediction accuracy similarity between models}. We observe that dCKA improves correlations, in terms of Spearman's $\rho$ and Kendall's $\tau$, with prediction accuracy on OOD test sets.}
\vskip 0.1in
\label{tab:OOD result}
\centering
\footnotesize
\begin{tabular}{ccccc}
\toprule
\multirow{2}{*}{Corruption level} & \multicolumn{2}{c}{CKA} & \multicolumn{2}{c}{dCKA} \\ \cline{2-5} 
                       & $\rho$        & $\tau$        & $\rho$         & $\tau$        \\ \hline
1                      & \makecell{0.147\\(0.004)}      & \makecell{0.103\\(0.003)}      & \makecell{0.151\\(0.004)}       & \makecell{0.105\\(0.003)}      \\
2                      & \makecell{0.150\\(0.004)}      & \makecell{0.106\\(0.003)}      & \makecell{0.157\\(0.004)}       & \makecell{0.110\\(0.003)}      \\
3                      & \makecell{0.132\\(0.004)}      & \makecell{0.094\\(0.002)}      & \makecell{\textbf{0.140}\\\textbf{(0.003)}}       & \makecell{\textbf{0.099}\\\textbf{(0.003)}}      \\
4                      & \makecell{0.130\\(0.003)}      & \makecell{0.091\\(0.002)}      & \makecell{\textbf{0.138}\\\textbf{(0.003)}}       & \makecell{\textbf{0.096}\\\textbf{(0.003)}}      \\
5                      & \makecell{0.135\\(0.003)}      & \makecell{0.094\\(0.002)}      & \makecell{0.140\\(0.004)}       & \makecell{0.098\\(0.003)}      \\ \hline
Average                & \makecell{0.139\\(0.002)}      & \makecell{0.098\\(0.001)}      & \makecell{\textbf{0.145}\\\textbf{(0.002)}}       & \makecell{\textbf{0.102}\\\textbf{(0.001)}}      \\ \hline
ID accuracy                     & \makecell{0.163\\(0.020)}      & \makecell{0.116\\(0.014)}      & \makecell{0.167\\(0.020)}       & \makecell{0.118 \\(0.014)}      \\\bottomrule
\end{tabular}
\end{table}

\begin{figure}[t]
\begin{center}
\centerline{\includegraphics[width=\columnwidth]{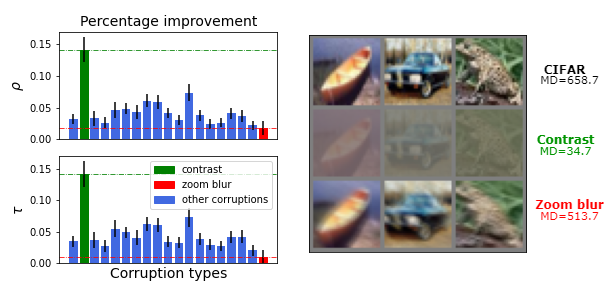}}
\caption{\textbf{dCKA improves CKA on each corruption type.} \textit{Left:} percentage improvement of dCKA over CKA and corresponding standard error on each type of corruptions. \textit{Right:} visualization of corruptions with largest improvement (`contrast', green) and smallest improvement (`zoom blur', red). We observe that `contrast' is more different from the uncorrupted dataset compared with `zoom blur', in terms of mean pair-wise distance (MD) of images.}
\label{fig: ood improvement hist}
\end{center}
\vskip -0.2in
\end{figure}

CIFAR-10-C dataset contains 19 different corruptions and 5 levels for each corruption. We average rank correlations over all blocks of ResNet-18 as \citet{ding2021grounding} because the ranking between similarity measures were shown to be consistent across different layers/blocks. We report the averaged rank correlation over all types of corruptions in Table \ref{tab:OOD result}. We observe that dCKA is more correlated with OOD accuracy on all 5 levels of corruption, especially on levels 3-5, compared with CKA. Moreover, we also notice marginal improvements of dCKA in terms of in distribution accuracy (ID accuracy). Figure \ref{fig: ood improvement hist} \textit{Left} shows the improvement of dCKA vs. CKA for each corruption averaged over 5 corruption levels, and we see that deconfounded CKA improves the most in the `contrast' corruption ($15\%$) and the least in the `zoom blur' corruption ($1\%$). We visualize examples of original in-distribution images together with the above two corruptions in Figure \ref{fig: ood improvement hist} \textit{Right}, and we observe that `contrast' is very different from the in-distribution `CIFAR' images whereas `zoom blur' is more similar to the original images visually. Moreover, we compute the mean pairwise Euclidean distances (MD) between images for each type. We find the MD of the original CIFAR test set (MD$=658.7$) is close to the `zoom blur' (MS=$513.7$), and very far from the `contrast' corruption (MD$=34.7$). Hence, the benefit of dCKA vs. CKA appear greatest when the OOD domain is least similar to the original domain, which aligns with the expectation that dCKA is less domain specific due to the correction for the input domain population structure.



\section{Discussion}
In this study, we investigated the confounding effect of the input similarity structure on commonly used similarity measures between neural networks representations. The confounder can lead to spuriously high similarity even for two completely random neural networks and counter-intuitive conclusions when cross-domain inputs have to be considered, for example in transfer learning and out-of-distribution generalization setting. 

We proposed a simple deconfounding algorithm by regressing out the input similarity structure from the representation similarity structure of each neural network. The deconfounded similarities studied in this paper retain the invariance properties of the original similarities, which are necessary when applied to understand neural network training. Moreover, deconfounding can significantly improve the consistency of similarities when inputs are from multiple domains. For instance, deconfounded similarities show more consistent results with domain similarities in both vision and language transfer learning tasks, and they are more correlated with out-of-distribution accuracy of neural networks trained with different initializations.

There are still few limitations and open questions. We assumed that the confounder is linearly separable from the representation similarity in Eq.\ref{eq: linear regression}, and showed that adding higher-order polynomial terms cannot improve the model evidence in Figure \ref{fig: optimal order}. However, it is still possible that the input similarity structure is not entirely \textit{additively} separable from representation similarity structures, especially for deeper layers, and this may explain the fact that deconfounding similarities are more beneficial for shallow layers (e.g., Table \ref{tab: null-distribution} and Figure \ref{fig: noisy NNs across domain}). One possible solution is regressing out the similarity structure in the previous layer instead of the input layer, which improves detecting similar networks from random networks (i.e., Table \ref{tab: null-distribution}) for deep blocks (Appendix \ref{app: detect similar NNs}). However, this discards information from all previous layers and eventually loses the ability of representing the similarity between functional behaviors. We consider this as an open question to motivate progress on developing more consistent similarity measures.
\newpage
\bibliographystyle{icml2022}

\newpage
\appendix
\onecolumn
\section{Training details of NNs}
\label{sec: training details}
\subsection{ResNet training on CIFAR-10}
We trained 50 ResNet-18 models without any pretrained model from different initializations on the CIFAR-10 training set. We use 200 epochs with batch size 128. We train models with SGD: learning rate 0.1, momentum 0.9, and weight decay 5e-4. We also use cosine annealing learning rate with $\text{T}_{max}=200$.

The averaged accuracy of trained models on CIFAR-10 test set is $0.89$, with standard deviation $0.3\%$.

\subsection{ResNet training on DomainNet}

\begin{figure}[ht]
  \centering
  \includegraphics[width=0.8\linewidth]{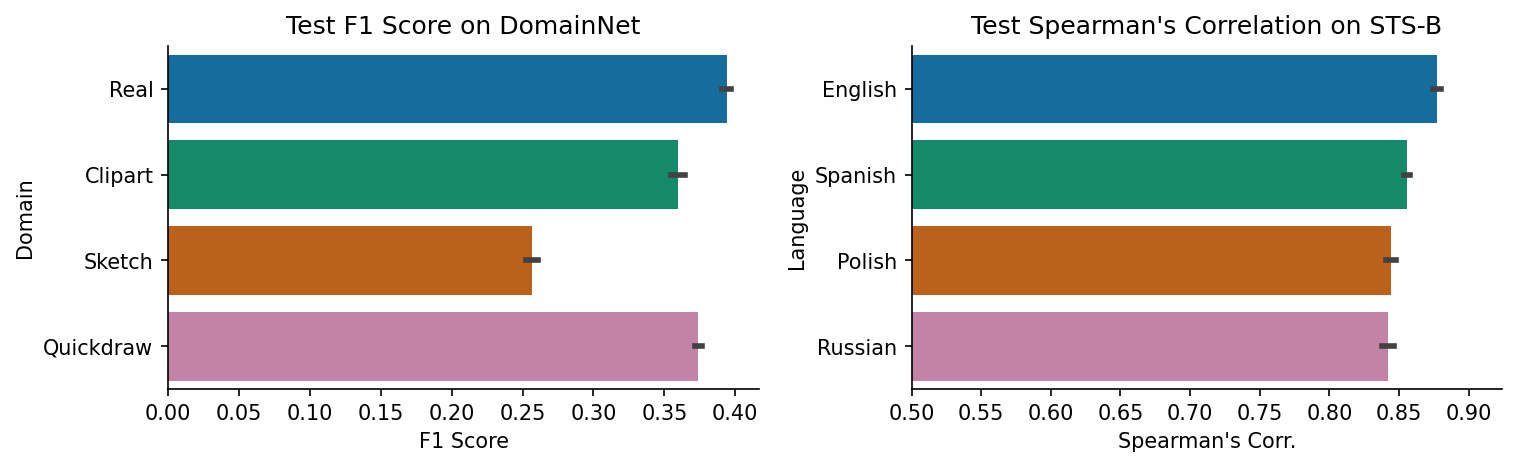}
  \caption{\emph{Left} Test F1 scores for each of the domains from DomainNet for Resnet-50 models. We can see that the Real domain, which is the most similar to Imagenet achieves the highest F1 score. \emph{Right} Test Spearman's correlation for the XLM-RoBERTa model when finetuned on each language.}
  \label{fig-app:transfer-acc}
\end{figure}

We finetune Resnet-50 models which had been pretrained on the Imagenet dataset. Separate models are trained on each domain of the DomainNet dataset, namely Real, Clipart, Sketch and Quickdraw. The DomainNet task is to classify each image among 345 different classes. For each domain, we repeat the training for 10 random restarts. To keep the training uniform, we sample 5000 images from each domain and train the Resnet model for 2000 iterations with a batch size of 32. All the input images are resized to $224 \times 224$ pixels, and we perform no other form of data augmentation. We use the AdamW optimizer \citep{loshchilov2018decoupled} with a base learning rate of $1e-3$, which was varied using a cosine annealing scheduler with a warm-up of 600 steps. Figure \ref{fig-app:transfer-acc} \emph{Left} shows the F1 scores of the finetuned model for each domain. As expected, the score for the Real domain is the highest since it is the most similar to Imagenet. The higher score on the Quickdraw domain is possibly due to the simplicity of the Quickdraw input distribution (since it consists of doodles) as compared to the other domains.

\subsection{XLM-RoBERTa training on STS-B}
Since we are training on different languages, we use the XLM-RoBERTa model as the base model and finetune it on different STS-B tasks on English, Spanish, Polish and Russian languages. For finetuning we use  AdamW optimizer with a learning rate of $2e-5$ which is linearly annealed, with 30\% of total steps for warmup, for 3 epochs. We use a batch size of 8 and regularize the training with a weight decay of $0.01$. Figure \ref{fig-app:transfer-acc} \emph{Right} shows Spearman's correlation of the predicted sentence similarity with the ground truth. The performance on English is the highest while that on Russian is the lowest. This performance across the languages correlates perfectly with the domain similarity reported in Section \ref{sec:domain-shift}.

\section{Durbin–Watson tests for noise correlation.}
\label{app: DW test}
In Eq.\ref{eq: linear regression}, we consider the collection of distance between each pair of examples is the dataset of the linear regression. A potential issue is that the noise term $\boldsymbol{\epsilon}^{m}_{f}$ can be correlated for different pairs, while we usually assume independent noise in model Eq.\ref{eq: linear regression}. Because the noise $\epsilon_{f, ij}^{m}$ of the distance between example $i$ and $j$ and $\epsilon_{f, ik}^{m}$ of the distance between example $i$ and $k$ contain the same information about example $i$, which might induce correlation between $\epsilon_{f, ij}^{m}$ and $\epsilon_{f, ik}^{m}$. Checking noise correlation is important to justify if the model is misspecified.

For each $i$, we apply the Durbin-Watson (DW) test on $\{\epsilon_{f, ij}^{m}|\forall j\}$, and average the test statistics over the index $i$. DW is always in $[0,4]$, and if there is no evidence of noise correlation, the test statistics equals to 2. Otherwise, the closer to 0 the statistic, the more evidence for positive correlation, while the closer to 4 means a negative correlation.

\begin{figure}[ht]
  \centering
  \includegraphics[width=0.5\linewidth]{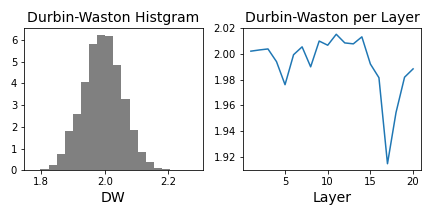}
  \caption{Durbin-Watson histogram and averaged DW statistics for each layer.}
  \label{fig app: DW results}
\end{figure}

We show the histogram of DW statistics as well as the averaged DW for each layer in Figure \ref{fig app: DW results}, where no serious noise correlation is observed.

\section{Proofs of propositions}
\label{app: proof}
\subsection{Proof of Proposition \ref{proposition: orthogonal transformation invariance}}
If the similarity measure $k(\cdot,\cdot)$ is invariant to orthogonal transformation, the representational similarity matrices are invariant to orthogonal transformation too. Thus, for any two full-rank orthonormal matrices $U$ and $V$, such that $U^{T}U=I$ and $V^{T}V = I$, we have:
\begin{equation}
\begin{split}
K(X^{m_1}_{{f_1}}U, X^{m_1}_{{f_1}}U) = K(X^{m_1}_{{f_1}}, X^{m_1}_{{f_1}});\;\; K(X^{m_2}_{{f_2}}V, X^{m_2}_{{f_2}}V) = K(X^{m_2}_{{f_2}}, X^{m_2}_{{f_2}}),
\end{split}
\end{equation}
and the deconfounded RSMs are also invariant to orthogonal transformation:
\begin{equation}
\begin{split}
&dK(X^{m_1}_{{f_1}}U, X^{m_1}_{{f_1}}U) = K(X^{m_1}_{{f_1}}U, X^{m_1}_{{f_1}}U)-\hat{\alpha}^{m_1}_{f_1}K^{0}= K(X^{m_1}_{{f_1}}, X^{m_1}_{{f_1}})-\hat{\alpha}^{m_1}_{f_1}K^{0}=dK(X^{m_1}_{{f_1}}, X^{m_1}_{{f_1}});\\
&dK(X^{m_2}_{{f_2}}V, X^{m_2}_{{f_2}}V) = K(X^{m_2}_{{f_2}}U, X^{m_2}_{{f_2}}U)-\hat{\alpha}^{m_2}_{f_2}K^{0}= K(X^{m_2}_{{f_2}}, X^{m_2}_{{f_2}})-\hat{\alpha}^{m_2}_{f_2}K^{0}=dK(X^{m_2}_{{f_2}}, X^{m_2}_{{f_2}}).
\end{split}
\end{equation}
Therefore the deconfounded similarity is invariant to orthogonal transformation:
\begin{equation}
s(dK(X^{m_1}_{{f_1}}U, X^{m_1}_{{f_1}}U),dK(X^{m_2}_{{f_2}}V, X^{m_2}_{{f_2}}V))=s(dK(X^{m_1}_{{f_1}}, X^{m_1}_{{f_1}}), dK(X^{m_2}_{{f_2}}, X^{m_2}_{{f_2}})),
\end{equation}
which completes the proof.

Note that the PSD approximation will not affect the orthogonal invariance property, because deconfounded RSMs are invariant to orthogonal transformations.

\subsection{Proof of Proposition \ref{proposition: isotropic scaling invariance}}
We use dRSA as an example to show the proof. For any $\gamma,\theta\in\mathbb{R}$, we have:
\begin{equation}
\begin{split}
K(\gamma X^{m_1}_{{f_1}}, \gamma X^{m_1}_{{f_1}}) = \gamma^2K(X^{m_1}_{{f_1}}, X^{m_1}_{{f_1}});\;\; K(\theta X^{m_2}_{{f_2}}, \theta X^{m_2}_{{f_2}}) = \theta^2K(X^{m_2}_{{f_2}}, X^{m_2}_{{f_2}}),
\end{split}
\end{equation}
because of using the Euclidean distance. Moreover, the deconfounded RSMs are scaled with $\lambda^2$ and $\theta^2$ as well, because the regression coefficient, $\alpha$ in Eq.\ref{eq: linear regression coef}, is scaled with the same value. Therefore, the rank correlation between two scaled deconfounded RSMs does not change because the rank is invariant to scaling all objects.

The PSD approximation will not affect the isotropic scaling invariance property too, because the PSD approximation matrix will be scaled at the same level.
\newpage

\section{Recursive deconfounded similarity on detecting similar networks from random networks.}
\label{app: detect similar NNs}
From Table \ref{tab: null-distribution}, we observe that although deconfounded similarities improve detecting semantically similar neural networks from random NNs, no similarity can identify ImageNet-CIFAR pairs for deep layers. One hypothesis is that the confounding effect of input similarity cannot be approximated well with additively separable functions for deeper layers, because of the model nonlinearity added by each neural network layer.

Here we consider a natural extension of deconfounding input similarity: instead of regressing out the input similarity directly (in Eq.\ref{eq: deconfound similarity structure}), we regress out the representation similarity structure from the previous layer recursively:
\begin{equation}
\begin{split}
&dK^{m_1}_{{f_1}}=K^{m_1}_{{f_1}}-\hat{\beta}^{m_1}_{f_1}K^{m_1-1}_{{f_1}};\\
&dK^{m_2}_{{f_2}}=K^{m_2}_{{f_2}}-\hat{\beta}^{m_2}_{f_2}K^{m_2-1}_{{f_2}}.
\label{eq: recursive deconfound similarity structure}
\end{split}
\end{equation}
Although Eq.\ref{eq: recursive deconfound similarity structure} has the same additively separable assumption as Eq.\ref{eq: deconfound similarity structure}, but Eq.\ref{eq: recursive deconfound similarity structure} is easier to satisfy because it only assumes additively separable for one layer. We call the similarity generated by Eq.\ref{eq: recursive deconfound similarity structure} recursive deconfounded similarity.

\begin{table}[ht]
\caption{Proportion of ImageNet-CIFAR ResNets pairs identified from random ResNets.}
\vskip 0.1in
\label{tab app: null-distribution}
\centering
\footnotesize
\begin{tabular}{ccccc}
\toprule
Block       & dCKA   & rdCKA     & dRSA       & rdRSA      \\ \hline
 1       & 1.0, 1.0 & 1.0, 1.0       & 1.0, 1.0  & 1.0, 1.0     \\
 2     & 1.0, 1.0  & 1.0, 1.0        & 1.0, 1.0  & 0.44, 0.98     \\
 3      & 1.0, 1.0 & 1.0, 1.0            & 0.0.0, 0.08 & 1.0, 1.0     \\
 4     & 1.0, 1.0 & 0.42, 1.0          & 0.0, 0.0   & 0.08, 0.02      \\
 5       & 0.0, 0.04 & 0.0, 0.78        & 0.0, 0.0  & 0.18, 0.0       \\
 6     & 0.0, 0.02 & 0.0, 0.04           & 0.0, 0.0   & 0.28, 0.4      \\
 7      & 0.0, 0.0  & 0.0, 0.0          & 0.0, 0.0    & 0.0, 0.0      \\
 8      & 0.0, 0.0  & 0.18, 0.12          & 0.0, 0.0    & 0.02, 0.18     \\ \hline
Average  & \textbf{0.5}, 0.51& 0.45, \textbf{0.62}  & 0.25, 0.26 & \textbf{0.38, 0.45}\\ \bottomrule
\end{tabular}
\end{table}

\begin{figure}[ht]
  \centering
  \includegraphics[width=0.8\linewidth]{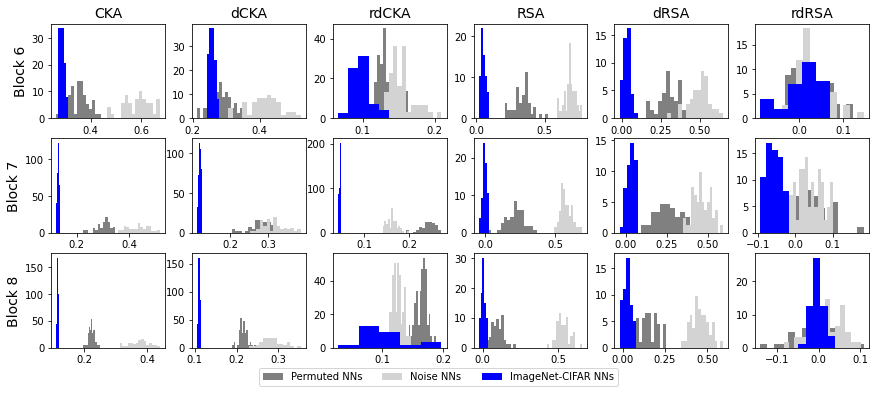}
  \caption{Histogram of each similarity measure for the last three blocks.}
  \label{fig app: similarity null distribution}
\end{figure}

We apply recursive deconfounded similarity, such as rdCKA and rdRSA, on the experiments of detecting similar networks from random networks described in Section \ref{sec: detect similar networks}. We show the comparison results between dCKA/dRSA with rdCKA/rdRSA in Table \ref{tab app: null-distribution}, where we observe a marginal improvement from dCKA to rdCKA but a significant improvement from dRSA to rdRSA. However, the proportion of identified similar networks is still low for deeper layers. Thus, we consider that ImageNet and CIFAR-10 learn different high-level representations because they contain different classes of images, as mentioned in the main text.

We visualize the histogram of each similarity measure for the last 3 blocks in Figure \ref{fig app: similarity null distribution}. We observe that CKA/dCKA and RSA/dRSA are much smaller than two null distributions, while rdCKA/rdRSA can have a similar level as the corresponding null distribution.

\section{Consistency of NNs in-domain similarities.}
\label{app: consistency in-domain}
In Section \ref{sec: ood consistency test}, we test the consistency of different NN similarities across 19 different domains. In this section, we verify if NN similarities are consistent across different input samples from the same domain. We repeat the same procedure described in Section \ref{sec: ood consistency test}, except that we calculate the similarity $s(f_i,f^{*})$ on 20 different sets of examples sampled from the same domain, i.e., the CIFAR-10 test set, instead of 19 different OOD domains.

\begin{figure}[ht]
  \centering
  \includegraphics[width=0.5\linewidth]{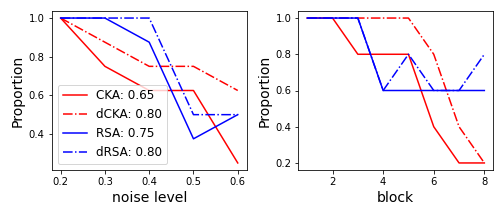}
  \caption{Proportion of identified similar NNs across different samples from the same domain. We observe that the deconfounded similarity can still identify more similar models compared with the corresponding original similarity.}
  \label{fig app: consistency with id domain}
\end{figure}

We show the results in Figure \ref{fig app: consistency with id domain}, where we observe that deconfounding can improve CKA/RSA with different inputs from the same domain, but the improvement is marginal compared with cross-domain experiments: $23\%$ for CKA (from 0.65 to 0.8) and $7\%$ for RSA (from 0.75 to 0.8), although the proportion of identified similar NNs are much larger than the cross-domain examples for all similarity measures.


\end{document}